\documentclass[10pt,conference]{IEEEtran}

\IEEEoverridecommandlockouts

\usepackage{cite}
\usepackage{amsmath,amssymb,amsfonts}
\usepackage{algorithm}
\usepackage{algorithmic}
\usepackage{graphicx}
\usepackage{textcomp}
\usepackage[table,xcdraw]{xcolor}
\usepackage{booktabs}
\usepackage{lscape}
\usepackage{url}
\usepackage{multirow}
\usepackage[most]{tcolorbox}

\def\BibTeX{{\rm B\kern-.05em{\sc i\kern-.025em b}\kern-.08em
    T\kern-.1667em\lower.7ex\hbox{E}\kern-.125emX}}

\newtcolorbox{observationbox}{
    colback=gray!15,
    colframe=gray!15,
    boxrule=0pt,
    arc=2mm,
    left=6pt,
    right=6pt,
    top=6pt,
    bottom=6pt,
    width=\linewidth,
    before skip=8pt,
    after skip=8pt,
}

\newcommand{\method}{CodeShrink}
\newcommand{\bfr}{Blank-Free Rendering (BFR)}
\newcommand{\iap}{Dominant Token Selection (DTS)}
\newcommand{\aps}{Adaptive Compression Configuration (ACC)}

\begin{document}

\title{\method: Adaptive Visual Compression for Efficient Multimodal Code Understanding}

\author{
\IEEEauthorblockN{
Wenxin Tang\textsuperscript{1},
Jingyu Xiao\textsuperscript{2},
Zhenyu Liu\textsuperscript{3},
Zipeng Xie\textsuperscript{4},
Junliang Liu\textsuperscript{2},
Wang Luo\textsuperscript{5},\\
Yuan Jiang\textsuperscript{6},
Yintong Huo\textsuperscript{7},
Michael Lyu\textsuperscript{2}
}

\IEEEauthorblockA{
\textsuperscript{1}Tsinghua University
\textsuperscript{2}The Chinese University of Hong Kong
\textsuperscript{3}Northwestern Polytechnical University\\
\textsuperscript{4}Xi'an Jiaotong University
\textsuperscript{5}Sun Yat-sen University
\textsuperscript{6}Harbin Institute of Technology\\
\textsuperscript{7}Singapore Management University\\
\texttt{twx24@mails.tsinghua.edu.cn},
\texttt{whalexiao99@gmail.com}
}
}

\maketitle

\begin{abstract}
Rendering source code as images offers a promising way to reduce the input costs of Multimodal Large Language Models (MLLMs). Adjusting image resolution can trade visual token cost against content fidelity. However, resolution scaling alone overlooks two sources of inefficiency: blank regions created by line breaks and indentation, and code regions irrelevant to the current instruction. Moreover, the best compression setting varies across inputs, tasks, and models, limiting fixed-ratio strategies. We propose CodeShrink, an adaptive visual compression framework with three components. Blank-Free Rendering replaces whitespace-dependent layouts with compact layouts and explicit structural markers, removing layout-induced tokens. Adaptive Compression Configuration uses a lightweight agent trained with reinforcement learning to predict a per-input setting that balances token efficiency and readability. Dominant Token Selection jointly analyzes the instruction and code image to prune task-irrelevant visual tokens during inference. We evaluate CodeShrink on code question answering, clone detection, and code completion. CodeShrink reduces visual token use by up to 71.2\% while matching or exceeding uncompressed text-only inputs, and consistently outperforms text-based and visual compression baselines across all three tasks. These results show that combining layout compaction, adaptive configuration, and instruction-aware pruning can make multimodal code understanding more efficient. Our code is available at \url{https://github.com/vinsontang1/CodeShrink}.
\end{abstract}


\section{Introduction}
Large Language Models (LLMs) have become increasingly important in automated software engineering, demonstrating strong capabilities across a wide range of code intelligence tasks, such as code question answering~\cite{shi2026codeocr,rando2025longcodebench}, code clone detection~\cite{alam2023gptclonebench}, and code completion~\cite{guo2023longcoder} and code generation~\cite{CoderEval, RepoGenix, A3CodeGen, xiao2026visualrepair}. By modeling source code as natural-language-like textual sequences, modern LLMs have demonstrated strong capabilities in reasoning over program semantics, supporting developers in understanding, maintaining, and generating code. However, source code often contains long contexts, repetitive syntax, and structurally sparse formatting, making token consumption a critical bottleneck~\cite{natural,DietCode} in code intelligence. As model inference costs and context length scale with the number of input tokens, token compression has become an important direction for scalable code intelligence.

To reduce input costs, prior studies such as LLMLingua~\cite{jiang2023llmlingua}, LongLLMLingua~\cite{jiang2024longllmlingua}, LLMLingua-2~\cite{pan2024llmlingua2}, and LongCodeZip~\cite{shi2025LongCodeZip} have attempted to compress code contexts directly in the textual token space. However, text-based compression remain constrained by the text modality itself and may introduce explicit semantic loss. In contrast, the image modality offers a key advantage in compressibility~\cite{wei2025deepseek}: image resolution can be continuously adjusted to scale token costs without altering the underlying content. Motivated by this, recent work such as CodeOCR~\cite{shi2026codeocr} has explored an alternative paradigm: rendering source code into images and leveraging Multimodal Large Language Models (MLLMs) for visual code understanding, offering a more computationally efficient alternative to text representations.



Rendering code as images introduces several key design choices that critically affect both token efficiency and performance. First, code images are inherently sparser than text: rendering source code with original line breaks and indentation often introduces substantial blank regions on the code image, which are encoded as visual tokens alongside code characters yet carry little semantic value. Second, image resolution directly governs visual token cost: while lower resolutions reduce token consumption, they inevitably compress character size, code structure, and syntax highlighting, degrading code readability for MLLMs. Hence, two natural questions arise: \textbf{(1) Are all visual tokens on the code image equally necessary? (2) How should resolution be determined to balance token efficiency and code readability?}

\begin{figure}[t]
    \centering
  \includegraphics[width=0.48\textwidth]{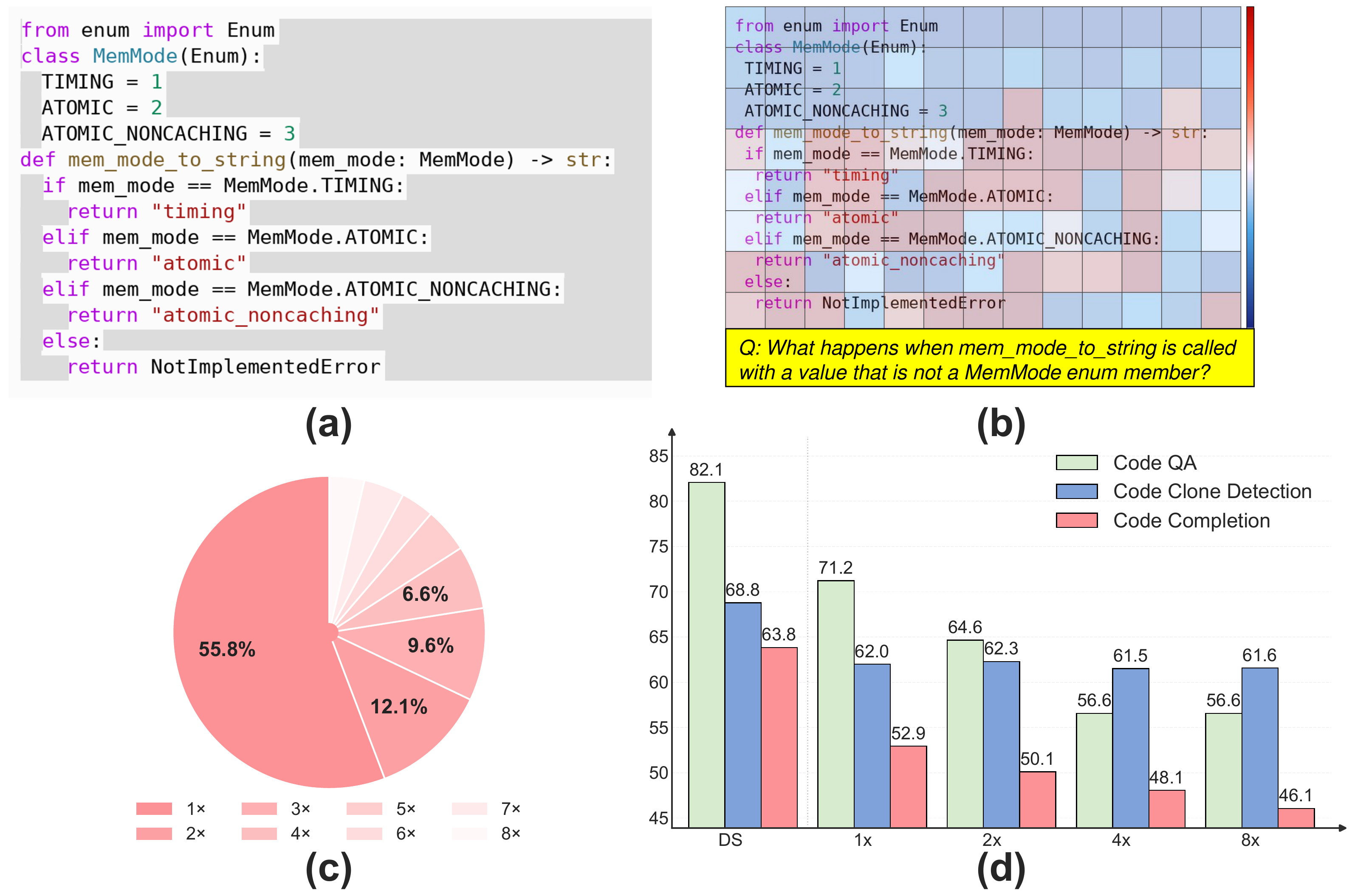}
  \caption{\textbf{Preliminary study of code image compression.}
(a) Original code rendering contains substantial blank regions, shown in gray.
(b) Attention visualization shows that task-relevant code regions receive higher attention than other regions.
(c) The preferred scale ratio varies across code completion samples.
(d) Dynamic selection (DS) surpasses the best fixed scale ratio across all three tasks. Code QA and code clone detection are evaluated by accuracy, and code completion is evaluated by edit similarity (ES).}
\label{fig:pre}
  \label{fig:pre}
\end{figure}
To answer these questions, we first identify \textit{visual redundancy} in code images (Section~\ref{sec:visual-token-redundancy}) by analyzing rendered code images and MLLM attention distributions over visual tokens. Our layout and attention visualizations reveal that redundant visual tokens originate from uninformative regions, such as blank areas caused by line breaks and indentation, and code-character regions weakly related to the downstream tasks.  To examine the influence of \textit{scale ratios} (Section~\ref{sec:compression-preference}), we further evaluate code intelligence tasks under different scale ratios and compare fixed-ratio strategies with an oracle dynamic selection strategy. The analysis indicates that the optimal trade-off between performance and token scale ratio varies across downstream tasks. These findings demonstrate that effective code-image compression requires not only eliminating visual redundancy, but also adaptively determining the optimal resolution to accommodate varying task demands.



Based on these findings, we propose \method, an adaptive visual compression framework for MLLM-based code understanding that eliminates redundancy at three stages: removing formatting-induced blank tokens at the rendering stage, pruning task-irrelevant visual tokens at the inference stage, and adaptively selecting the optimal resolution via a learned compression configuration. First, we propose \bfr, which decouples code structure from whitespace by replacing verbose layouts with compact representations and explicit structural markers, thereby preserving indentation and line-break semantics while eliminating layout-induced blank visual tokens. Second, we design \iap, which selects important visual tokens by analyzing the instruction and code image, enabling the MLLM to focus on task-relevant regions. Third, we introduce an \aps \ strategy, which employs reinforcement learning to dynamically select the compression configuration for different tasks. Together, these components advance \method \ beyond static resolution adjustment toward task-aware and model-adaptive visual token optimization.

We evaluate \method \ on multiple code-related tasks, including code question answering, code clone detection, and code completion. Experimental results show that \method \ can substantially reduce visual token cost while maintaining or even improving downstream task performance. For instance, on Python code question answering, \method\ reaches 82.3\% accuracy and surpasses the 81.0\% of the uncompressed text-only input under about 40\% visual token reduction, and it consistently outperforms both text-based and visual compression baselines. These results demonstrate that, through denser code image representation, dynamic token pruning and resolution selection, \method\ achieves substantial token compression with near-lossless task performance. Our contributions are as follows:
\begin{itemize}
    \item We conduct an in-depth analysis of visual token representation in MLLM-based code understanding, identifying two visual redundancy in the typical rendering of code as images.
    \item We propose \method, an adaptive compression framework that reduces layout redundancy through blank-free rendering, removes task-irrelevant visual tokens through dominant token selection, and learns a dynamic resolution and compression ratio selection strategy to adapt different tasks through reinforcement learning.
    \item We conduct extensive experiments across multiple code understanding tasks, demonstrating that \method \ achieves substantial visual token compression while maintaining or even improving task performance.
\end{itemize}

\section{Background}

\subsection{Visual-based Code Understanding}

Visual-based code understanding treats source code as a rendered image rather than a textual token sequence, enabling MLLMs to perform code intelligence tasks directly from visual inputs. Let $s$ denote a source code snippet and $q$ a task instruction. A renderer $R$ maps $s$ to a code image $I = R(s)$. Given $I$ and $q$, a Multimodal Large Language Model $M$ produces the output $y = M(I, q)$, where $y$ is the task-specific prediction,for example, in code clone detection $y$ is a binary label indicating whether two snippets are semantically equivalent. Unlike text-based code understanding, where $s$ is tokenized into a textual sequence, the visual paradigm presents the whole snippet as a single rendered image, so the model reads program content directly from pixels, which shifts the textual input into a compressible visual input.

\subsection{MLLMs' Process Pipeline for Code Intelligence}

Visual code understanding from code images typically have four stages: a renderer, a visual encoder, a modality projector, and an LLM. The renderer $R$ converts code text into an image $I$ under a chosen resolution and code image layout. The visual encoder splits $I$ into patches and encodes them into a sequence of visual tokens. The projector aligns these visual tokens with the LLM embedding space, and the LLM then consumes the visual tokens together with the instruction tokens to generate the output. The computational cost of the MLLM grows with its input sequence length, which can be written as $n = n_{img} + n_{text}$,
where $n_{img}$ is the number of visual tokens from the code image and $n_{text}$ is the number of instruction tokens.  In code understanding tasks, $n_{text}$ corresponds to a fixed-length task instruction and remains roughly constant across samples, while $n_{img}$ scales with the length and complexity of the source code, dominating the total input and constituting the primary target for compression.


\section{Preliminary Study}
\subsection{Visual Token Redundancy in Code Images}
\label{sec:visual-token-redundancy}
In practical code understanding, MLLMs do not need to attend equally to every visual region in a code image. Some regions in a code image come from formatting rather than program semantics. As shown in Fig.~\ref{fig:pre}(a), when original line breaks and indentation are directly preserved on the code image, large blank regions appear around code blocks. These blank regions contain no code characters, but still occupy image area and are converted into visual tokens by the vision encoder. Under a limited resolution or visual token budget, the renderer often has to reduce the font size to fit the complete code, making code characters harder for the model to recognize under high scale ratios.

Beyond explicit blank regions, code-character regions also show uneven importance. To analyze how MLLMs attend to code images, we compute attention scores of visual tokens on a code question answering example. As shown in Fig.~\ref{fig:pre}(b), code regions related to the question receive clearly higher attention scores, while some task-irrelevant code regions have scores close to background regions. This indicates that even within regions containing code characters, some visual tokens contribute little to the current instruction.
\begin{observationbox}
\textbf{Observation 1:} Existing code images contain visual token redundancy from blank and task-irrelevant code regions.
\end{observationbox}

\subsection{Compression Configuration Preferences Vary Across Samples}
\label{sec:compression-preference}

Existing code-as-image methods usually predefine a scale ratio and apply the same setting to all samples. However, different samples may require different trade-offs between visual clarity and token cost. To analyze this variation, we use code completion as an example and identify, for each sample, the highest scale ratio that preserves its task score. As shown in Fig.~\ref{fig:pre}(c), the most cost-effective ratios are distributed across multiple ratio values rather than dominated by a single fixed setting. \textbf{This suggests that one scale ratio cannot fit all samples.}
\begin{figure*}[t]
  \centering
  \includegraphics[width=0.95\textwidth]{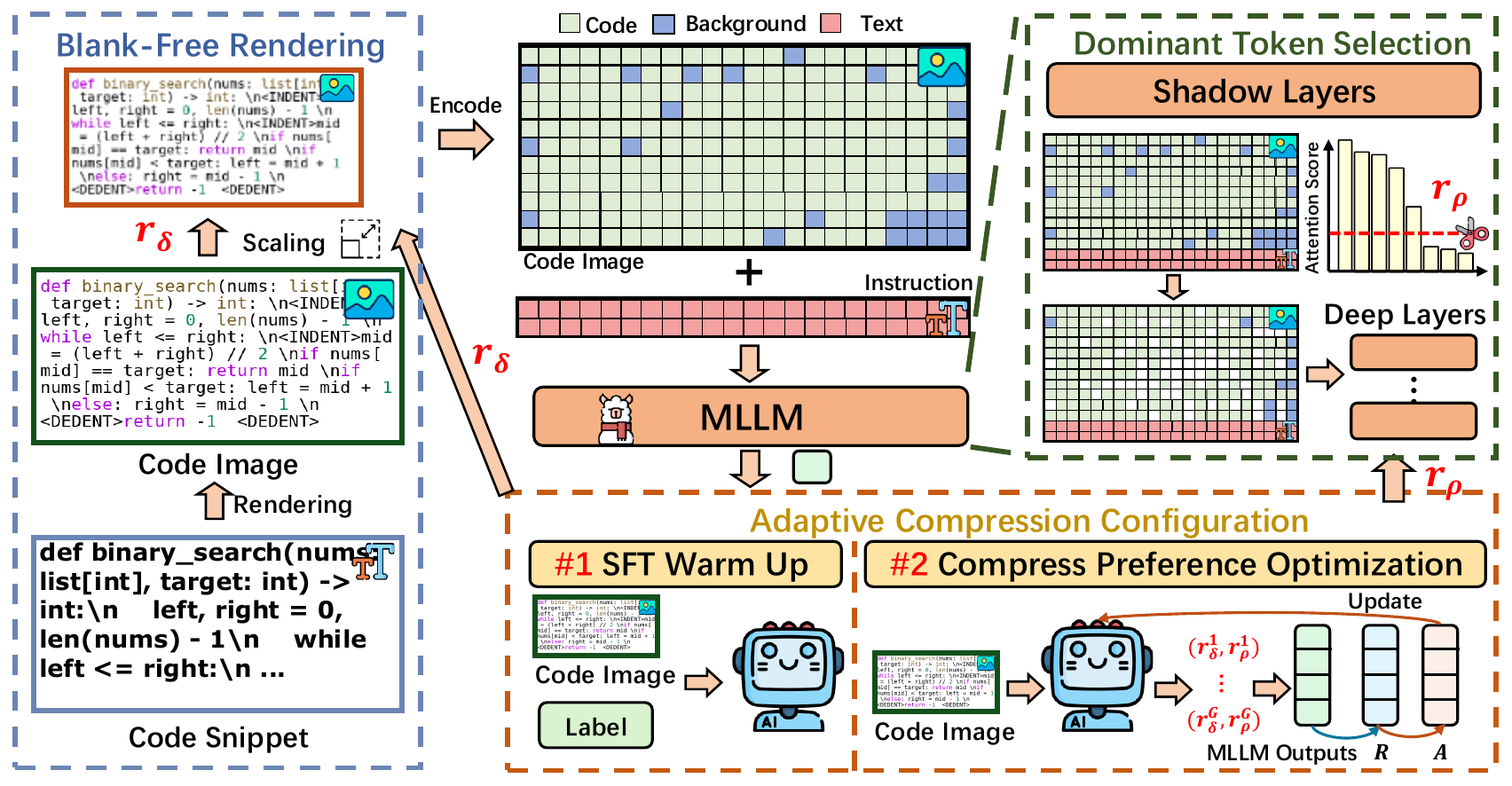}
  \caption{The framework of \method.}
  \label{fig:overview}
\end{figure*}
We further compare fixed scale ratios with an oracle dynamic selection strategy. For the dynamic strategy, each sample selects the best-performing result from all candidate ratios. As shown in Fig.~\ref{fig:pre}(d), dynamic selection outperforms the best fixed ratio across multiple tasks. The improvement is especially clear on code question answering and code completion, where the scores increase from 71.3\% and 52.9\% to 82.1\% and 63.8\%, respectively. These results indicate that dynamic selection can exceed the performance upper bound of any single fixed ratio. 


\begin{observationbox}

\textbf{Observation 2:} The optimal trade-off between visual clarity and token cost varies across samples and tasks.
\end{observationbox}

\section{Methodology}

\textbf{Overview.} Based on the two observations, we propose \method\ (as shown in Fig.~\ref{fig:overview}) to \textit{adaptively} compress visual tokens for visual-based code understanding. \method\ first applies the \bfr\ module (Section~\ref{sec:bfr}) to generate dense code images by flattening code with explicit structural markers. Then the \iap\ module (Section~\ref{sec:dts}) selects task-relevant visual tokens during inference while pruning tokens with low informational value. Finally, the \aps\ strategy (Section~\ref{sec:ars}) learns compression configuration preferences from downstream task feedback and the given model, dynamically optimizing compression configuration for each sample.

\renewcommand{\algorithmiccomment}[1]{\hfill// #1}

\subsection{Blank-Free Rendering}
\label{sec:bfr}

\begin{algorithm}[t]
\caption{Blank-Free Rendering}
\label{alg:bfr}
\begin{algorithmic}[1]
\REQUIRE code snippet $s$, scale ratio $r_{\delta}$, patch size $\tau$, aspect ratio $\alpha$
\ENSURE rendered code image pages $\mathcal{I}$
\STATE $T \leftarrow \textsc{Linearize}(s)$
\STATE $B \leftarrow \lceil n_s / r_\delta \rceil$ \COMMENT{Target visual-token budget}
\STATE $p \leftarrow 1$
\REPEAT
    \STATE $(W,H) \leftarrow \textsc{Canvas}(B / p,\ \alpha)$
    \STATE $f \leftarrow \max\{\, f : \textsc{Pages}(T,W,H,f) \le p \,\}$ \COMMENT{Binary search}
    \STATE $p' \leftarrow \textsc{Pages}(T,W,H,f)$
    \IF{$p' \le p$}
        \STATE \textbf{break}
    \ELSE
        \STATE $p \leftarrow p'$ \COMMENT{Raise page count}
    \ENDIF
\UNTIL{converged}
\STATE $H \leftarrow \tau \cdot \lceil H_{\text{used}} / \tau \rceil$ \COMMENT{Remove blank patch overhead}
\STATE $\{L_1,\dots,L_p\} \leftarrow \textsc{Partition}(\textsc{Layout}(T,W,f),\ p)$
\STATE $\mathcal{I} \leftarrow \textsc{Draw}(\{L_k\},\ W,H,f)$
\RETURN $\mathcal{I}$
\end{algorithmic}
\end{algorithm}

Code images contain blank regions from formatting like line breaks and indentation. We design \bfr \ to render code at high density within a fixed visual-token budget, allocating tokens to semantically-meaningful code content. As shown in Algorithm~\ref{alg:bfr}, given a code snippet $s$, we first perform structural linearization (line~1): we strip the whitespace of each line, drop empty lines, and encode the two-dimensional layout into a one-dimensional token stream using \texttt{<INDENT>} and \texttt{<DEDENT>} to represent indentation and de-indentation, together with a visible newline marker to preserve structure. The scale ratio $r_{\delta}$, provided by the agent in Section~\ref{sec:ars}, then fixes a visual-token budget $B=\lceil n_s/r_\delta \rceil$ (line~2), where $n_s$ is the number of tokens in $s$.

The core objective of the layout algorithm is to maximize font size for readability while keeping the total visual token count within the budget and ensuring all code content fits within the image. We cast this as
\begin{equation}
f^\star=\max\big\{\, f : \textsc{Pages}(T,W,H,f)\le p \,\big\},
\end{equation}
and since the required page count $\textsc{Pages}$ is non-decreasing in the font size, the largest feasible font is found exactly by binary search (line~6), where $p$ denotes the current number of pages, initialized to $1$ (line~3). The code image shape is determined from the current budget $B/p$ and a target aspect ratio $\alpha$, subject to the minimum and maximum pixel resolutions accepted by the model (line~5).
When the smallest font cannot fit the code within the current page count, we raise the page count to the required value and retry. This fixed-point iteration converges because the page count is monotonically increasing. Finally, we align the code image height up to the patch grid to remove the blank-patch overhead of a trailing partial row (line~14), distribute the code lines evenly across pages, and draw them. In this way, the visual tokens previously occupied by blanks are removed, compacting the code representation.

\subsection{Dominant Token Selection}
\label{sec:dts}

After BFR, visual redundancy still remains within both code regions and blank regions. Motivated by the observation in Section~\ref{sec:visual-token-redundancy}, we apply region-aware pruning to foreground code tokens and background blank tokens separately. The key idea is to decide the most important visual tokens relevant to the task information after visual and instruction information have been fused.. As shown in Fig.~\ref{fig:overview}, the compressed image is first split into visual tokens by the MLLM visual encoder. We then use edge and color-gradient cues to partition these tokens into foreground code tokens (green) and background blank tokens (blue). These visual tokens are fed together with the instruction tokens (red) into the shallow MLLM layers, allowing token importance to be estimated under the current instruction.

Specifically, the \aps\ module in Section~\ref{sec:ars} provides the pruning ratios $r_\rho=(r_{\rho_f},r_{\rho_b})$ for foreground and background tokens. After the visual and instruction tokens are fused in the shallow MLLM layers, we use the query vector $q$ of the final instruction token and the key vector $k_i$ of each visual token to measure token importance. Averaging over $H$ attention heads, the importance score of visual token $i$ is computed as:
\begin{equation}
a_i=\frac{1}{H}\sum_{h=1}^{H}
\operatorname{softmax}\!\left(
\frac{q^{(h)}{k_i^{(h)}}^{\top}}{\sqrt{d}}
\right).
\end{equation}
We rank tokens by importance within the foreground and background sets separately. The lowest $r_{\rho_f}$ fraction of foreground tokens and the lowest $r_{\rho_b}$ fraction of background tokens are removed, while the retained visual tokens continue through the remaining layers together with the text tokens. Because pruning physically removes visual tokens and their KV-cache entries once during prefill, subsequent layers operate on a shorter sequence and the effective $n_{img}$ is reduced. Since the importance scores are conditioned on the instruction, the same code image can preserve different tokens for different tasks, guiding the model toward instruction-relevant code regions.

\subsection{Adaptive Compression Configuration}
\label{sec:ars}

Both BFR and DTS are controlled by the ratios $r_\delta$ and $r_\rho$, whose optimal values vary across samples and tasks, as established in Section~\ref{sec:compression-preference}. A single fixed setting is therefore suboptimal. To address this, we introduce a Config Agent that predicts a per-sample compression configuration through reinforcement learning (RL).

The Config Agent $\pi_\theta$ is a tiny MLLM. Given the $1\times$ rendered image from BFR, it outputs a configuration $a=(r_\delta, r_\rho)$, where the scale ratio $r_\delta$ sets the rendering budget $B=\lceil n_s/r_\delta\rceil$ in Section~\ref{sec:bfr} and $r_\rho=(r_{\rho_f}, r_{\rho_b})$ are the foreground and background pruning ratios in Section~\ref{sec:dts}. Since the Config Agent only needs to capture the downstream model's visual preference, rather than to solve the downstream task itself, we expose only the first page of the rendered image to the Agent. This keeps training efficient and avoids noise from task-specific reasoning. We define the action space as a grid:
\begin{equation}
\mathcal{A}=\mathcal{R}_\delta\times\mathcal{R}_{\rho_f}\times\mathcal{R}_{\rho_b},
\end{equation}
where $\mathcal{R}_\delta$, $\mathcal{R}_{\rho_f}$, and $\mathcal{R}_{\rho_b}$ are the discrete candidate sets of the scale ratio and of the foreground and background pruning ratios.
To avoid repeatedly querying the MLLM for the same rollout during training, we evaluate every configuration $a\in\mathcal{A}$ on the sampled instances in advance, producing a precomputed environment $\mathcal{D}=\mathcal{D}_{sft}\cup\mathcal{D}_{rl}$, whose entries $(I,a,C(a),y(a))$ record the compression $C(a)$ and prediction correctness $y(a)$ of each configuration $a$ on an instance $I$. The disjoint subsets $\mathcal{D}_{sft}$ and $\mathcal{D}_{rl}$ serve the supervised fine-tuning (SFT) and reinforcement learning (RL) stages respectively. For a configuration $a$ with budget split $B=B_f+B_b$, the retained visual-token count after pruning is $N(a)=(1-r_{\rho_f})B_f+(1-r_{\rho_b})B_b$, and we measure compression relative to the $1\times$ image as
\begin{equation}
C(a)=\frac{n_s}{N(a)},
\end{equation}
where a larger $C(a)$ indicates fewer retained tokens. We further denote by $y(a)\in\{0,1\}$ whether the MLLM answers correctly under $a$.

Training proceeds in two stages. The SFT warm-up provides a sensible initialization: for each instance in $\mathcal{D}_{sft}$, among the configurations that the MLLM answers correctly, we select the one with the largest compression $C(a)$ as the label, and supervise the Agent to predict it. This anchors the Agent at an aggressive yet still-legible operating point. We then refine the Agent with Group Relative Policy Optimization (GRPO)~\cite{shao2024deepseekmath} so that it also learns to avoid configurations that break legibility. The reward retains the compression gain only when the answer stays correct and is normalized to $[0,1]$,
\begin{equation}
R(a)=\frac{C(a)}{\max_{a'\in\mathcal{A}}C(a')}\,y(a).
\end{equation}
For each instance in $\mathcal{D}_{rl}$, we draw a group of $G$ configurations $\{a_1,\dots,a_G\}$ and compute the group-relative advantage
\begin{equation}
\hat{A}_i=\frac{R(a_i)-\operatorname{mean}(\{R(a_j)\}_{j=1}^{G})}{\operatorname{std}(\{R(a_j)\}_{j=1}^{G})}.
\end{equation}
Denoting the importance ratio by $\eta_i=\pi_\theta(a_i)/\pi_{\theta_{\text{old}}}(a_i)$, the per-sample clipped surrogate is
\begin{equation}
\ell_i(\theta)=\min\!\big(\eta_i\hat{A}_i,\ \operatorname{clip}(\eta_i,1-\epsilon,1+\epsilon)\,\hat{A}_i\big),
\end{equation}
and the Agent maximizes:
\begin{equation}
\mathcal{J}(\theta)=\mathbb{E}\!\left[\frac{1}{G}\sum_{i=1}^{G}\ell_i(\theta)\right]-\beta\,D_{KL}\!\left(\pi_\theta\,\|\,\pi_{\text{ref}}\right).
\end{equation}
The reward couples compression with correctness, allowing the Agent to learn aggressive compression within legibility constraints and remove the need for any hand-tuned ratio.

Additionally, to enhance the downstream MLLM's OCR ability on code images, we few-shot fine-tune it.

\begin{table*}[t]
\centering
\caption{Performance comparison of different compression methods on the Python benchmark across code clone detection, code question answering, and code completion. \textbf{Bold values} indicate the optimal performance, while \underline{underlined values} indicate the second-best performance.}
\label{tab:python_main}
\resizebox{0.75\textwidth}{!}{%
\begin{tabular}{lcccccccc}
\hline
\multicolumn{1}{l|}{\multirow{2}{*}{\textbf{Method}}} &
  \multicolumn{3}{c|}{\textbf{Code Clone Detection (\%)}} &
  \multicolumn{2}{c|}{\textbf{Code QA (\%)}} &
  \multicolumn{3}{c}{\textbf{Code Completion (\%)}} \\ \cline{2-9}
\multicolumn{1}{l|}{} &
  \textbf{Acc} &
  \textbf{F1} &
  \multicolumn{1}{c|}{\textbf{R}} &
  \textbf{Acc} &
  \multicolumn{1}{c|}{\textbf{R}} &
  \textbf{ES} &
  \textbf{EM} &
  \textbf{R} \\ \hline

\multicolumn{9}{c}{\textbf{Reference Baselines}} \\ \hline
\multicolumn{1}{l|}{Text-only}     & \underline{64.5} & \underline{46.0} & \multicolumn{1}{c|}{-}    & \underline{81.0} & \multicolumn{1}{c|}{-}    & 55.1  & \underline{27.0} & -    \\
\multicolumn{1}{l|}{NoCtx}    & -     & -     & \multicolumn{1}{c|}{-}    & 54.3 & \multicolumn{1}{c|}{-}    & 49.0  & 15.0 & -    \\ \hline

\multicolumn{9}{c}{\textbf{Text-based Token Compression}} \\ \hline
\multicolumn{1}{l|}{LLMLingua~\cite{jiang2023llmlingua}}     & 59.0 & 30.5 & \multicolumn{1}{c|}{25.1}  & 61.3 & \multicolumn{1}{c|}{25.5} & 54.1  & 24.0 & 20.8 \\
\multicolumn{1}{l|}{LongLLMLingua~\cite{jiang2024longllmlingua}} & 57.0 & 24.6 & \multicolumn{1}{c|}{23.4}  & 61.7 & \multicolumn{1}{c|}{25.9} & 53.8  & 22.0 & 21.2 \\
\multicolumn{1}{l|}{LLMLingua-2~\cite{pan2024llmlingua2}}    & 50.5 & 3.9  & \multicolumn{1}{c|}{\underline{29.7}} & 60.0 & \multicolumn{1}{c|}{35.6} & 50.2  & 14.0 & \underline{28.7} \\
\multicolumn{1}{l|}{LongCodeZip~\cite{shi2025LongCodeZip}}   & 60.5 & 34.7 & \multicolumn{1}{c|}{20.6}  & 62.7 & \multicolumn{1}{c|}{\underline{39.3}} & \underline{56.7}  & \underline{27.0} & \textbf{30.2} \\ \hline

\multicolumn{9}{c}{\textbf{Visual Token Compression}} \\ \hline
\multicolumn{1}{l|}{VisionZip~\cite{yang2025visionzip}}     & 50.0 & 29.9 & \multicolumn{1}{c|}{16.7} & \underline{65.7} & \multicolumn{1}{c|}{18.7} & 43.9  & 13.0  & 15.3 \\
\multicolumn{1}{l|}{FastV~\cite{chen2024fastv}}         & 53.5 & 13.5 & \multicolumn{1}{c|}{18.4} & 66.0 & \multicolumn{1}{c|}{22.1} & 52.0  & 17.0 & 18.0 \\ \hline

\multicolumn{9}{c}{\textbf{Code-as-Image}} \\ \hline
\multicolumn{1}{l|}{CodeOCR~\cite{shi2026codeocr}}       & 61.5 & 39.4 & \multicolumn{1}{c|}{-}    & 51.7 & \multicolumn{1}{c|}{-}    & 46.4  & 16.5 & -    \\
\multicolumn{1}{l|}{CodeShrink (Ours)}    & \textbf{68.0} & \textbf{65.9} & \multicolumn{1}{c|}{\textbf{41.4}} &
  \textbf{82.3} & \multicolumn{1}{c|}{\textbf{39.7}} & \textbf{60.3} &
  \textbf{27.5} & 21.1 \\ \hline
\end{tabular}%
}
\end{table*}

\begin{table}[h]
\centering
\caption{Performance comparison of different compression methods on the Java benchmark across code clone detection and code completion. \textbf{Bold values} indicate the optimal performance, while \underline{underlined values} indicate the second-best performance.}
\label{tab:java_main}
\resizebox{0.48\textwidth}{!}{%
\begin{tabular}{lcccccc}
\hline
\multicolumn{1}{l|}{\multirow{2}{*}{\textbf{Method}}} &
  \multicolumn{3}{c|}{\textbf{Code Clone Detection (\%)}} &
  \multicolumn{3}{c}{\textbf{Code Completion (\%)}} \\ \cline{2-7} 
\multicolumn{1}{c|}{} &
  \textbf{Acc} &
  \textbf{F1} &
  \multicolumn{1}{c|}{\textbf{R}} &
  \textbf{ES} &
  \textbf{EM} &
  \textbf{R} \\ \hline
\multicolumn{7}{c}{\textbf{Reference Baselines}} \\ \hline
\multicolumn{1}{l|}{Text-only}     & \underline{69.0} & \underline{55.6} & \multicolumn{1}{c|}{-}    & \underline{58.1} & 34.0 & -    \\
\multicolumn{1}{l|}{NoCtx}         & -     & -     & \multicolumn{1}{c|}{-}    & 48.8 & 14.5 & -    \\ \hline
\multicolumn{7}{c}{\textbf{Text-based Token Compression}} \\ \hline
\multicolumn{1}{l|}{LLMLingua~\cite{jiang2023llmlingua}}     & 60.5 & 36.1 & \multicolumn{1}{c|}{25.3}  & 57.3 & 25.5 & 28.1 \\
\multicolumn{1}{l|}{LongLLMLingua~\cite{jiang2024longllmlingua}} & 59.0 & 33.3 & \multicolumn{1}{c|}{35.8}  & \underline{58.1} & \underline{27.5} & 23.2 \\
\multicolumn{1}{l|}{LLMLingua-2~\cite{pan2024llmlingua2}}    & 58.5 & 29.1 & \multicolumn{1}{c|}{32.9} & 57.0 & 20.5 & 25.9 \\
\multicolumn{1}{l|}{LongCodeZip~\cite{shi2025LongCodeZip}}   & 64.0 & 45.0 & \multicolumn{1}{c|}{\underline{37.6}}  & 48.0 & 17.5 & 13.7 \\ \hline
\multicolumn{7}{c}{\textbf{Visual Token Compression}} \\ \hline
\multicolumn{1}{l|}{VisionZip~\cite{yang2025visionzip}}     & 60.5 & 34.7 & \multicolumn{1}{c|}{19.7} & 43.3 & 7.0  & 23.8 \\
\multicolumn{1}{l|}{FastV~\cite{chen2024fastv}}         & 59.0 & 30.5 & \multicolumn{1}{c|}{22.1} & 52.3 & 18.5 & \underline{28.5} \\ \hline
\multicolumn{7}{c}{\textbf{Code-as-Image}} \\ \hline
\multicolumn{1}{l|}{CodeOCR~\cite{shi2026codeocr}}       & 63.0 & 41.3 & \multicolumn{1}{c|}{-}  & 37.1 & 10.0 & -  \\
\multicolumn{1}{l|}{CodeShrink (Ours)} &
  \textbf{71.0} &
  \textbf{70.1} &
  \multicolumn{1}{c|}{\textbf{71.2}} &
  \textbf{63.8} &
  \textbf{34.0} &
  \textbf{32.4} \\ \hline
\end{tabular}%
}
\end{table}

\section{Experiments}

\subsection{Experiment Setup}
\subsubsection{Benchmark and Metrics}
We evaluate \method\ on three representative code understanding tasks. \textbf{Code Question Answering}, which requires the model to answer natural-language questions grounded in a given code snippet, is evaluated on the QA dataset proposed by CodeOCR~\cite{shi2026codeocr}, from which we randomly sample 300 instances and report accuracy. \textbf{Code Clone Detection}, which asks the model to judge whether two snippets are semantically equivalent, is built upon GPTCloneBench~\cite{alam2023gptclonebench}, from which we randomly sample 200 balanced Type-4 instances and report accuracy together with F1 Score. \textbf{Code Completion}, which requires the model to complete an unfinished snippet, is drawn from LongCodeCompletion~\cite{guo2023longcoder} with 200 randomly sampled instances and is measured by exact match (EM), which counts the proportion of predictions that exactly match the reference, and edit similarity (ES), which measures the character-level closeness between the prediction and the reference based on edit distance. 

Beyond task performance, we further quantify efficiency through the token reduction rate $R = 1-n_{img}^{c}/n_{img}$, where $n_{img}^{c}$ is the number of visual tokens after compression and $n_{img}$ is that of the original rendering, which is set to match the token count of the plain-text version of the code.

\subsubsection{Baselines} We compare \method\ against four groups of baselines. 

\textit{Reference Baselines} establish the performance bounds. \textbf{Text-Only} feeds the full source code as text without compression. \textbf{NoCtx} answers without access to any code. 

\textit{Text-based Token Compression} methods compress code in the textual space. \textbf{LLMLingua}~\cite{jiang2023llmlingua} is a prompt compression method that uses a budget controller and token-level iterative compression to shorten prompts while preserving the original semantics. \textbf{LLMLingua-2}~\cite{pan2024llmlingua2} is a task-agnostic method that casts compression as a token-level preserve-or-discard classification problem trained on distilled data. \textbf{LongLLMLingua}~\cite{jiang2024longllmlingua} is a long-context-oriented method that performs question-aware coarse-to-fine compression with document reordering to retain the context most relevant to the question. \textbf{LongCodeZip}~\cite{shi2025LongCodeZip} is a code-specific method that first ranks function-level chunks by instruction relevance and then selects finer-grained code blocks under an adaptive token budget. 

\textit{Visual Token Compression} methods prune visual tokens during inference. \textbf{FastV}~\cite{chen2024fastv} removesimage tokens according to attention after a designated LLM layer. \textbf{VisionZip}~\cite{yang2025visionzip} first selects dominant visual tokens via visual-encoder attention and then merges the remaining ones by similarity. 

\textit{Code-as-Image} methods render code into images. \textbf{CodeOCR}~\cite{shi2026codeocr} renders source code into images and controls visual token cost by adjusting the image resolution.

\subsubsection{Implementation Details} To train the Config Agent, we sample 100 instances per task and exhaustively evaluate the action space $\mathcal{A}=\mathcal{R}_\delta\times\mathcal{R}_{\rho_f}\times\mathcal{R}_{\rho_b}$, where the scale ratio takes $\mathcal{R}_\delta=\{1,2,\dots,8\}$, the foreground pruning ratio takes $\mathcal{R}_{\rho_f}=\{0,0.1,0.3,0.5\}$, and the background pruning ratio takes $\mathcal{R}_{\rho_b}=\{0.5,0.7,0.9,1\}$, yielding 38,400 configuration entries across the three tasks. These instances are further split into three disjoint parts, of which 50\% are used to fine-tune the OCR ability of the MLLM, 25\% for the SFT warm-up, and 25\% for the RL stage. As for the models, we adopt Qwen3.5-27B~\cite{team2026qwen3} as the downstream MLLM and Qwen3.5-0.8B~\cite{team2026qwen3} as the lightweight Config agent, both of which are fine-tuned with LoRA~\cite{hu2022lora} of rank 16 and a learning rate of 1e-4. The SFT warm-up runs for 1 epoch, while the GRPO stage runs for 2 epochs with a group size of $G=8$. To ensure a fair comparison across all methods, we adopt greedy decoding with temperature 0 and fix the random seed to 42 for reproducibility. Notably, benefiting from the less training samples, fine-tuning the MLLM takes only about 25 minutes on 8 NVIDIA H20 GPUs, while training the Config Agent takes about 40 minutes on a single H20 GPU.

\begin{table*}[t]
\centering
\caption{Cross-model comparison between vanilla rendering and our BFR rendering across three representative MLLMs, where for each pair the \colorbox[HTML]{DCEDD5}{green background} highlights the better result.}
\label{tab:image_resolution}
\resizebox{0.75\textwidth}{!}{%
\begin{tabular}{lcccccc}
\hline
\multicolumn{1}{l|}{\multirow{2}{*}{\textbf{Method}}} &
  \multicolumn{2}{c|}{\textbf{\begin{tabular}[c]{@{}c@{}}Code Clone Detection\\ (Acc/F1, \%)\end{tabular}}} &
  \multicolumn{2}{c|}{\textbf{\begin{tabular}[c]{@{}c@{}}Code QA\\ (Acc, \%)\end{tabular}}} &
  \multicolumn{2}{c}{\textbf{\begin{tabular}[c]{@{}c@{}}Code Completion\\ (ES/EM, \%)\end{tabular}}} \\ \cline{2-7}
\multicolumn{1}{l|}{} &
  \textbf{Vanilla} &
  \multicolumn{1}{c|}{\textbf{BFR}} &
  \textbf{Vanilla} &
  \multicolumn{1}{c|}{\textbf{BFR}} &
  \textbf{Vanilla} &
  \textbf{BFR} \\ \hline

\multicolumn{7}{c}{\textbf{Image (1x)}} \\ \hline
\multicolumn{1}{l|}{Qwen3-VL} &
  67.0/50.6 &
  \multicolumn{1}{c|}{\cellcolor[HTML]{DCEDD5}68.4/54.6} &
  58.5 &
  \multicolumn{1}{c|}{\cellcolor[HTML]{DCEDD5}63.2} &
  36.2/10.5 &
  \cellcolor[HTML]{DCEDD5}54.7/23.8 \\
\multicolumn{1}{l|}{GPT-5-mini} &
  \cellcolor[HTML]{DCEDD5}64.0/45.2 &
  \multicolumn{1}{c|}{62.6/41.9} &
  \cellcolor[HTML]{DCEDD5}76.4 &
  \multicolumn{1}{c|}{71.8} &
  49.1/21.9 &
  \cellcolor[HTML]{DCEDD5}51.4/19.6 \\
\multicolumn{1}{l|}{Gemini-3-Flash} &
  67.8/55.6 &
  \multicolumn{1}{c|}{\cellcolor[HTML]{DCEDD5}68.9/54.2} &
  76.7 &
  \multicolumn{1}{c|}{\cellcolor[HTML]{DCEDD5}77.1} &
  58.3/29.7 &
  \cellcolor[HTML]{DCEDD5}61.6/34.7 \\ \hline

\multicolumn{7}{c}{\textbf{Image (2x)}} \\ \hline
\multicolumn{1}{l|}{Qwen3-VL} &
  66.6/51.6 &
  \multicolumn{1}{c|}{\cellcolor[HTML]{DCEDD5}67.2/51.8} &
  49.7 &
  \multicolumn{1}{c|}{\cellcolor[HTML]{DCEDD5}56.6} &
  33.8/8.2 &
  \cellcolor[HTML]{DCEDD5}51.9/21.2 \\
\multicolumn{1}{l|}{GPT-5-mini} &
  \cellcolor[HTML]{DCEDD5}69.0/56.4 &
  \multicolumn{1}{c|}{65.7/46.4} &
  \cellcolor[HTML]{DCEDD5}75.1 &
  \multicolumn{1}{c|}{75.0} &
  47.1/17.7 &
  \cellcolor[HTML]{DCEDD5}48.3/16.1 \\
\multicolumn{1}{l|}{Gemini-3-Flash} &
  68.4/57.0 &
  \multicolumn{1}{c|}{\cellcolor[HTML]{DCEDD5}69.0/55.7} &
  76.4 &
  \multicolumn{1}{c|}{\cellcolor[HTML]{DCEDD5}76.5} &
  58.3/28.6 &
  \cellcolor[HTML]{DCEDD5}61.0/33.4 \\ \hline

\multicolumn{7}{c}{\textbf{Image (4x)}} \\ \hline
\multicolumn{1}{l|}{Qwen3-VL} &
  59.2/31.8 &
  \multicolumn{1}{c|}{\cellcolor[HTML]{DCEDD5}65.5/47.0} &
  50.6 &
  \multicolumn{1}{c|}{\cellcolor[HTML]{DCEDD5}57.4} &
  37.7/10.7 &
  \cellcolor[HTML]{DCEDD5}49.2/18.3 \\
\multicolumn{1}{l|}{GPT-5-mini} &
  \cellcolor[HTML]{DCEDD5}68.8/56.8 &
  \multicolumn{1}{c|}{66.1/49.6} &
  57.9 &
  \multicolumn{1}{c|}{\cellcolor[HTML]{DCEDD5}71.9} &
  44.2/14.9 &
  \cellcolor[HTML]{DCEDD5}46.8/12.0 \\
\multicolumn{1}{l|}{Gemini-3-Flash} &
  67.6/55.4 &
  \multicolumn{1}{c|}{\cellcolor[HTML]{DCEDD5}68.8/53.1} &
  77.2 &
  \multicolumn{1}{c|}{\cellcolor[HTML]{DCEDD5}78.7} &
  58.5/28.6 &
  \cellcolor[HTML]{DCEDD5}60.5/31.9 \\ \hline

\multicolumn{7}{c}{\textbf{Image (8x)}} \\ \hline
\multicolumn{1}{l|}{Qwen3-VL} &
  60.8/35.4 &
  \multicolumn{1}{c|}{\cellcolor[HTML]{DCEDD5}65.3/47.9} &
  51.3 &
  \multicolumn{1}{c|}{\cellcolor[HTML]{DCEDD5}58.6} &
  41.0/11.9 &
  \cellcolor[HTML]{DCEDD5}47.1/14.5 \\
\multicolumn{1}{l|}{GPT-5-mini} &
  64.0/47.0 &
  \multicolumn{1}{c|}{\cellcolor[HTML]{DCEDD5}65.6/46.5} &
  52.5 &
  \multicolumn{1}{c|}{\cellcolor[HTML]{DCEDD5}69.3} &
  44.1/13.5 &
  \cellcolor[HTML]{DCEDD5}45.7/12.2 \\
\multicolumn{1}{l|}{Gemini-3-Flash} &
  70.2/60.8 &
  \multicolumn{1}{c|}{\cellcolor[HTML]{DCEDD5}71.4/59.4} &
  76.3 &
  \multicolumn{1}{c|}{\cellcolor[HTML]{DCEDD5}77.0} &
  57.8/25.3 &
  \cellcolor[HTML]{DCEDD5}60.4/30.6 \\ \hline
\end{tabular}%
}
\end{table*}

\subsubsection{Research Questions} We conduct extensive experiments to answer the following research questions:
\begin{itemize}
\item \textbf{(RQ1) Overall Performance:} Can \method\ effectively compress visual tokens while preserving code understanding performance?
\item \textbf{(RQ2) Rendering Quality:} How does \bfr\ affect code readability across different models compared with vanilla code rendering?
\item \textbf{(RQ3) Ablation Study:} How does each component of \method\ contribute to the overall performance?
\item \textbf{(RQ4) ACC Deep Dive:} How do different tasks prefer different compression configurations, and how does this motivate \aps?
\item \textbf{(RQ5) Case Study:} Why does \method\ work?
\end{itemize}

\subsection{Overall Performance (RQ1)}

As shown in Table~\ref{tab:python_main} and Table~\ref{tab:java_main}, which report the main results on Python and Java, we can make the following observations: (1) \textbf{\method\ attains the best performance while substantially compressing visual tokens, even surpassing the uncompressed \textit{Text-only} input on most metrics.} On Python, it reaches 82.3\% accuracy on code question answering, which exceeds the 81.0\% of \textit{Text-only} under about 40\% token reduction. We attribute this to the synergy of the three modules, where BFR frees budget through dense rendering, ACC adaptively selects the configuration setting that best matches each sample, and \iap\ prunes instruction-irrelevant tokens at inference time, so that the limited budget is concentrated on task-relevant code and compression yields gains rather than degradation. (2) \textbf{\method\ achieves a consistently better compression-performance trade-off than both text-based and visual compression baselines.} Text-based methods such as the \textit{LLMLingua family} drop tokens directly in the textual modality, which tends to break code semantics and structure, as the F1 of \textit{LLMLingua2} collapses to merely 3.92\% on Python clone detection. Although visual compression instead operates at the image level and thereby avoids such modality damage, conventional methods still rely on task-agnostic fixed pruning and suffer large losses on code, where the EM of \textit{VisionZip} falls to only 13.0\% on Python code completion. As a reference, the \textit{CodeOCR} that renders code at a fixed 1$\times$ resolution lags clearly behind on demanding tasks such as code question answering, where it reaches merely 51.7\%, which further confirms the necessity of Adaptive Compression Configuration.

\begin{observationbox}
\textbf{Answer to RQ1:} \method\ delivers substantial visual token compression across all three code understanding tasks while maintaining or even exceeding the uncompressed input, and it consistently outperforms both text-based and visual compression methods.
\end{observationbox}

\begin{table*}[t]
\centering
\caption{Ablation study of \method\ on Python and Java, where each variant removes a single component while keeping the rest unchanged. \textit{w/o BFR}, \textit{w/o DTS}, \textit{w/o ACC}, and \textit{w/o SFT} respectively disable the blank-free rendering, the inference-time token pruning, the Config Agent, and the downstream fine-tuning. $R$ denotes the visual token reduction rate. \textbf{Bold values} indicate the optimal performance, while \underline{underlined values} indicate the second-best performance.}
\label{tab:ablation}
\resizebox{0.75\textwidth}{!}{%
\begin{tabular}{lcccccccc}
\hline
\multicolumn{1}{l|}{\multirow{2}{*}{\textbf{Method}}} &
  \multicolumn{3}{c|}{\textbf{Code Clone Detection (\%)}} &
  \multicolumn{2}{c|}{\textbf{Code QA (\%)}} &
  \multicolumn{3}{c}{\textbf{Code Completion (\%)}} \\ \cline{2-9}
\multicolumn{1}{l|}{} &
  \textbf{Acc} &
  \textbf{F1} &
  \multicolumn{1}{c|}{\textbf{R}} &
  \textbf{Acc} &
  \multicolumn{1}{c|}{\textbf{R}} &
  \textbf{ES} &
  \textbf{EM} &
  \textbf{R} \\ \hline
\multicolumn{9}{c}{\textbf{Python}} \\ \hline
\multicolumn{1}{l|}{CodeShrink} &
  \underline{66.5} & 63.4 & \multicolumn{1}{c|}{\textbf{41.0}} &
  \textbf{81.5} & \multicolumn{1}{c|}{\textbf{39.5}} &
  \textbf{59.9} & \textbf{27.0} & \textbf{20.9} \\
\multicolumn{1}{l|}{w/o BFR} &
  66.0 & \underline{63.9} & \multicolumn{1}{c|}{17.9} &
  78.5 & \multicolumn{1}{c|}{26.5} &
  58.4 & 26.0 & 14.2 \\
\multicolumn{1}{l|}{w/o DTS} &
  \textbf{67.0} & \textbf{64.1} & \multicolumn{1}{c|}{\underline{37.2}} &
  \underline{79.5} & \multicolumn{1}{c|}{33.8} &
  \underline{58.6} & \underline{26.5} & 17.9 \\
\multicolumn{1}{l|}{w/o ACC} &
  65.5 & 62.8 & \multicolumn{1}{c|}{16.6} &
  75.0 & \multicolumn{1}{c|}{19.1} &
  54.8 & 22.5 & 14.1 \\
\multicolumn{1}{l|}{w/o SFT} &
  66.0 & 62.7 & \multicolumn{1}{c|}{33.6} &
  78.5 & \multicolumn{1}{c|}{\underline{34.4}} &
  57.9 & 25.0 & \underline{18.8} \\ \hline
\multicolumn{9}{c}{\textbf{Java}} \\ \hline
\multicolumn{1}{l|}{CodeShrink} &
  \textbf{70.5} & \textbf{69.4} & \multicolumn{1}{c|}{\textbf{53.5}} &
  - & \multicolumn{1}{c|}{-} &
  \textbf{64.2} & \textbf{33.5} & \textbf{32.0} \\
\multicolumn{1}{l|}{w/o BFR} &
  \underline{69.0} & 67.5 & \multicolumn{1}{c|}{22.7} &
  - & \multicolumn{1}{c|}{-} &
  62.0 & 32.0 & 19.5 \\
\multicolumn{1}{l|}{w/o DTS} &
  \underline{69.0} & \underline{68.2} & \multicolumn{1}{c|}{\underline{45.7}} &
  - & \multicolumn{1}{c|}{-} &
  \underline{62.1} & \underline{32.5} & 26.7 \\
\multicolumn{1}{l|}{w/o ACC} &
  64.5 & 63.2 & \multicolumn{1}{c|}{22.5} &
  - & \multicolumn{1}{c|}{-} &
  56.2 & 28.5 & 27.4 \\
\multicolumn{1}{l|}{w/o SFT} &
  68.5 & 67.9 & \multicolumn{1}{c|}{35.1} &
  - & \multicolumn{1}{c|}{-} &
  60.9 & 31.5 & \underline{28.2} \\ \hline
\end{tabular}%
}
\end{table*}

\begin{figure*}[h]
  \centering
  \includegraphics[width=0.95\textwidth]{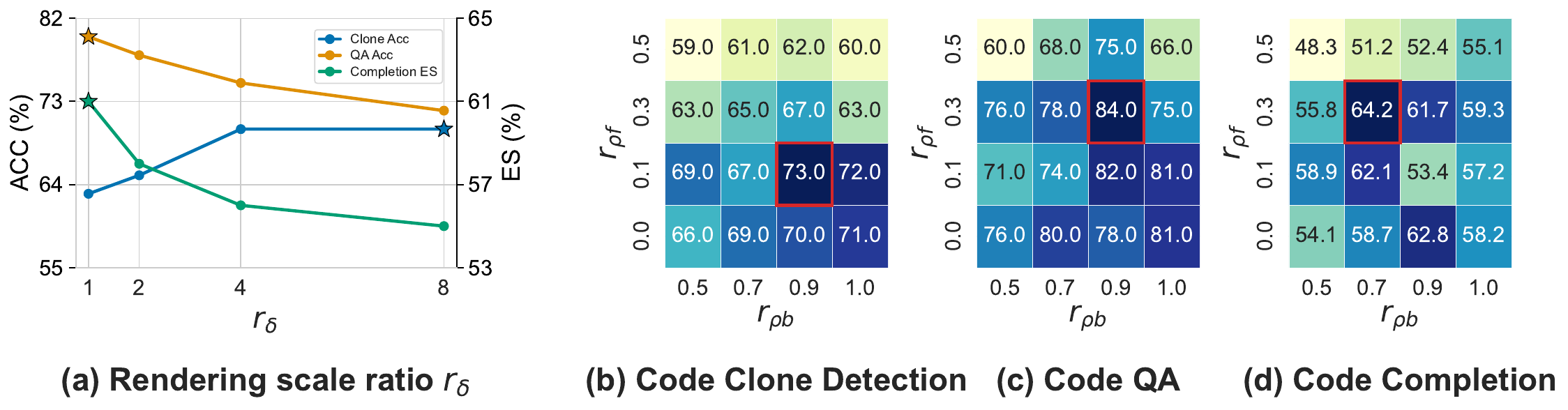}
  \caption{Task-dependent compression configuration of \method\ on the three code understanding tasks. (a) shows the effect of the scale ratio $r_\delta$, where the stars mark the best operating point of each task. (b--d) show the effect of the foreground and background pruning ratios $r_{\rho_f}$ and $r_{\rho_b}$ on code clone detection, code QA, and code completion respectively, where the red box marks the best configuration of each task.}

  \label{fig:param}
\end{figure*}

\subsection{Rendering Quality (RQ2)}

Table~\ref{tab:image_resolution} presents the cross-model comparison between vanilla rendering and our \bfr\ rendering, where only the rendering function is replaced while the rest of the pipeline is fixed, across three representative MLLMs, namely Qwen3-VL-235B-A22B-Instruct~\cite{qwen3technicalreport} (Qwen3-VL), GPT-5-mini~\cite{openai2025gpt5mini}, and Gemini-3-Flash~\cite{google2025gemini3flash} and resolutions ranging from 1$\times$ to 8$\times$. Several key observations emerge from this analysis. Following CodeOCR, we randomly sample 200 instances from each dataset and report results averaged over five independent runs. (1) \textbf{BFR rendering outperforms vanilla rendering on almost all models and resolution levels without any model-specific adaptation.} Since BFR operates only at the rendering stage and leaves the model untouched, it brings consistent gains even to the closed-source GPT-5-mini and Gemini-3-Flash that are never trained, which indicates that this rendering optimization is model-agnostic and can be transferred across different MLLMs as a plug-and-play improvement. (2) \textbf{The gains become especially pronounced on long code tasks and under high compression.} On the long code task, which is code completion, BFR rendering leads by a clear margin, where the ES and EM of Qwen3-VL at 1$\times$ rise from 36.2\% and 10.5\% to 54.7\% and 23.8\% respectively. As compression intensifies, vanilla rendering degrades markedly because of blank redundancy and the shrinking font size, whereas BFR preserves readability under aggressive compression by removing blanks and keeping the font legible, which further widens its advantage, as the code question answering accuracy of GPT-5-mini at 8$\times$ jumps from 52.5\% to 69.3\%. Although GPT-5-mini occasionally favors vanilla rendering at low compression, BFR rendering regains a clear lead once entering the high compression regime.

\begin{observationbox}
\textbf{Answer to RQ2:} As a model-agnostic and plug-and-play improvement, BFR rendering generally enhances code understanding across different MLLMs and compression levels, with its advantage being most pronounced on long code tasks and under high compression.
\end{observationbox}

\subsection{Ablation Study (RQ3)}


We conduct an ablation study by removing one component at a time. Specifically, \textit{w/o BFR} replaces Blank-Free Rendering with the vanilla rendering of CodeOCR, \textit{w/o DTS} disables visual token pruning, \textit{w/o ACC} replaces the Config Agent with a fixed compression configuration, and \textit{w/o SFT} removes downstream MLLM fine-tuning.

In Table~\ref{tab:ablation}, \textit{w/o ACC} causes the most pronounced degradation, where accuracy and token reduction drop simultaneously. The QA accuracy on Python falls from 81.5\% to 75.0\% and the clone detection accuracy on Java falls from 70.5\% to 64.5\%, while R drops from 53.5\% to 22.5\%, which confirms that per-sample adaptive configuration selection is the key to balancing performance and compression. BFR mainly contributes to the density of the code image. Removing it leaves accuracy largely unchanged but causes a sharp drop in R from 53.5\% to 22.7\% on Java clone detection, because without blank-free rendering the code becomes less dense and the font shrinks, which tends to distort the code under high scale ratios, so the Config Agent favors lower scale ratios to preserve performance. DTS and SFT provide mild but steady complementary gains, as removing DTS slightly lowers compression with almost no change in accuracy, and removing SFT leads to small declines in both accuracy and compression.

\begin{observationbox}
\textbf{Answer to RQ3:} Each component of \method\ has a positive impact on the results, and combining all components yields the best performance.
\end{observationbox}


\subsection{ACC Deep Dive (RQ4)}


To better understand ACC, we analyze the compression configuration preferences of different code understanding tasks. Specifically, we study the scale ratio $r_\delta$ of BFR and the foreground/background pruning ratios $r_{\rho_f}$ and $r_{\rho_b}$ of DTS. The analysis is conducted on 100 randomly sampled instances for each task.

\subsubsection{The scale ratio $r_\delta$} To examine its effect, we vary $r_\delta\in\{1,2,4,8\}$, where a larger value denotes stronger compression and lower resolution. As shown in Figure~\ref{fig:param}(a), the optimal scale ratio clearly differs across tasks. Specifically, QA and code completion peak at 1$\times$ and then decline steadily as compression increases, because both rely on fine-grained character and contextual details that a lower resolution erodes. In contrast, the accuracy of clone detection rises with compression and reaches its best at 8$\times$, improving from about 63\% to 70\%, since judging clones only requires structural similarity and thus moderate compression even helps the model focus on the overall structure.

\subsubsection{The pruning ratios $r_{\rho_f}$ and $r_{\rho_b}$} Similarly, to analyze pruning, we vary $r_{\rho_f}\in\{0.0,0.1,0.3,0.5\}$ and $r_{\rho_b}\in\{0.5,0.7,0.9,1.0\}$. As shown in Figure~\ref{fig:param}(b--d), the optimal foreground pruning ratio is likewise task-dependent. In particular, clone detection prefers a small ratio of 0.1, whereas QA and code completion favor 0.3. This gap arises because the code in clone detection is short, so each foreground token already carries enough information and overly aggressive pruning would discard key content. In contrast, the longer code in QA and completion produces foreground tokens that often contain meaningless fragments such as character tails, so both the foreground and the background can be pruned more aggressively, with a relatively high background ratio $r_{\rho_b}$ consistently benefiting all tasks since the background is dominated by low-information blank regions.

\subsubsection{Task-dependent optimal configuration} Taken together, the above results reveal that the optimal compression configuration is highly task-dependent. The best scale ratio shifts from 1$\times$ for QA and completion to 8$\times$ for clone detection, and the best foreground pruning ratio likewise varies from 0.1 to 0.3 across tasks, so no single configuration stays optimal for all of them. As a result, any manually fixed setting is inevitably a compromise that is suboptimal on part of the tasks, and the gap widens further once we note that different samples within the same task may also favor different configurations. This observation directly motivates our design, where instead of committing to one global setting, the Config Agent in our ACC module predicts a tailored configuration for each sample and thereby stays close to the per-task optimum.

\begin{observationbox}
\textbf{Answer to RQ4:} The optimal compression configuration of \method\ is not fixed but varies across tasks and samples, so hand-tuned fixed parameters can hardly remain optimal. Our Config Agent addresses this by adaptively selecting the configuration for each sample, achieving consistently strong performance across tasks.
\end{observationbox}

\subsection{Case Study (RQ5)}

\begin{figure*}[t]
  \centering
  \includegraphics[width=0.99\textwidth]{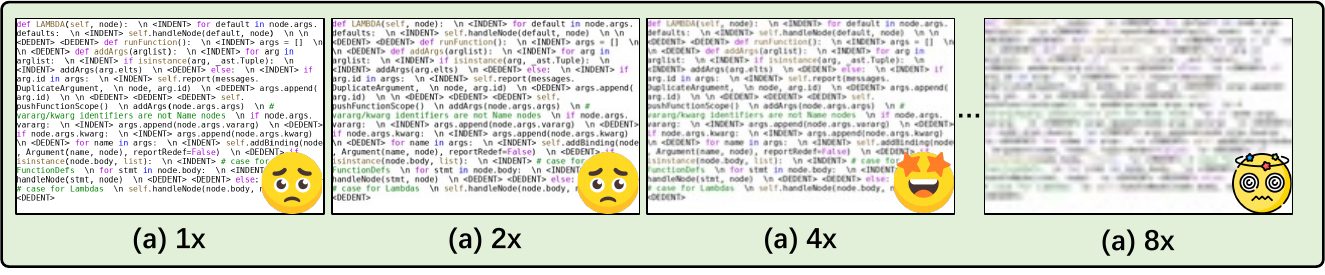}
  \caption{Case study of the configuration selection result. The Config Agent selects 4$\times$ for this sample, since 1$\times$ and 2$\times$ waste tokens on an unnecessarily high resolution while 8$\times$ makes the code illegible.}

  \label{fig:case_scale}
\end{figure*}

To explain why \method\ works, we present visualized cases that illustrate its three core steps: \bfr, \iap, and \aps.

\subsubsection{\bfr} 
\begin{figure}[h]
  \centering
  \includegraphics[width=0.48\textwidth]{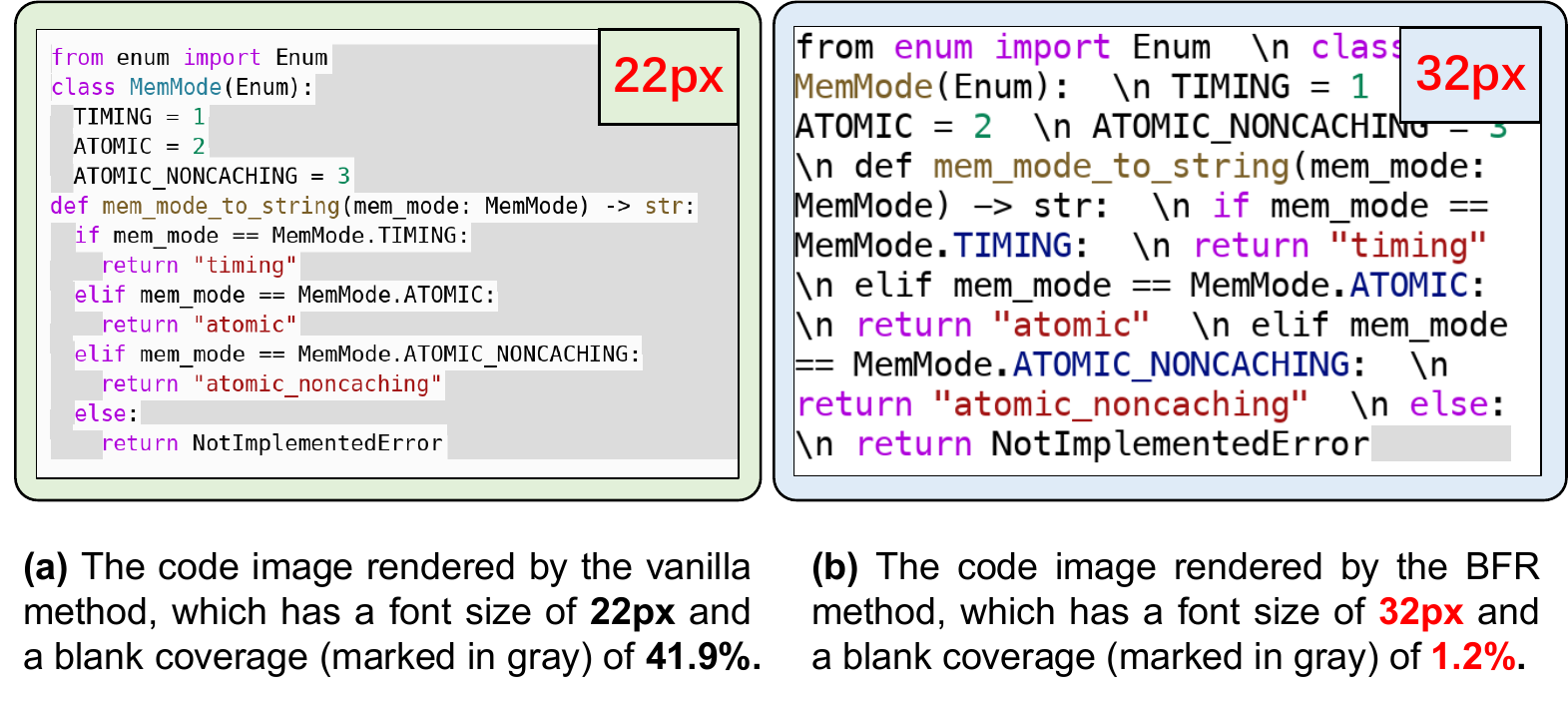}
  \caption{Code images rendered by different methods.}

  \label{fig:case_render}
\end{figure}

As shown in Fig.~\ref{fig:case_render}, the vanilla rendering preserves the original whitespace and indentation of the source code, so a large portion of the code image is occupied by blank regions that are still encoded as visual tokens yet carry no semantic content, leading to a blank coverage of 41.9\% (marked in gray) while the font size remains only 22px. In contrast, BFR keeps the structural information through compact markers and removes the redundant blanks, reducing the blank coverage to 1.2\%. Under the same token budget, this allows the font size to grow to 32px, which makes individual characters and the overall code structure considerably clearer and easier for the model to recognize.

\subsubsection{\iap} 
\begin{figure}[h]
  \centering
  \includegraphics[width=0.48\textwidth]{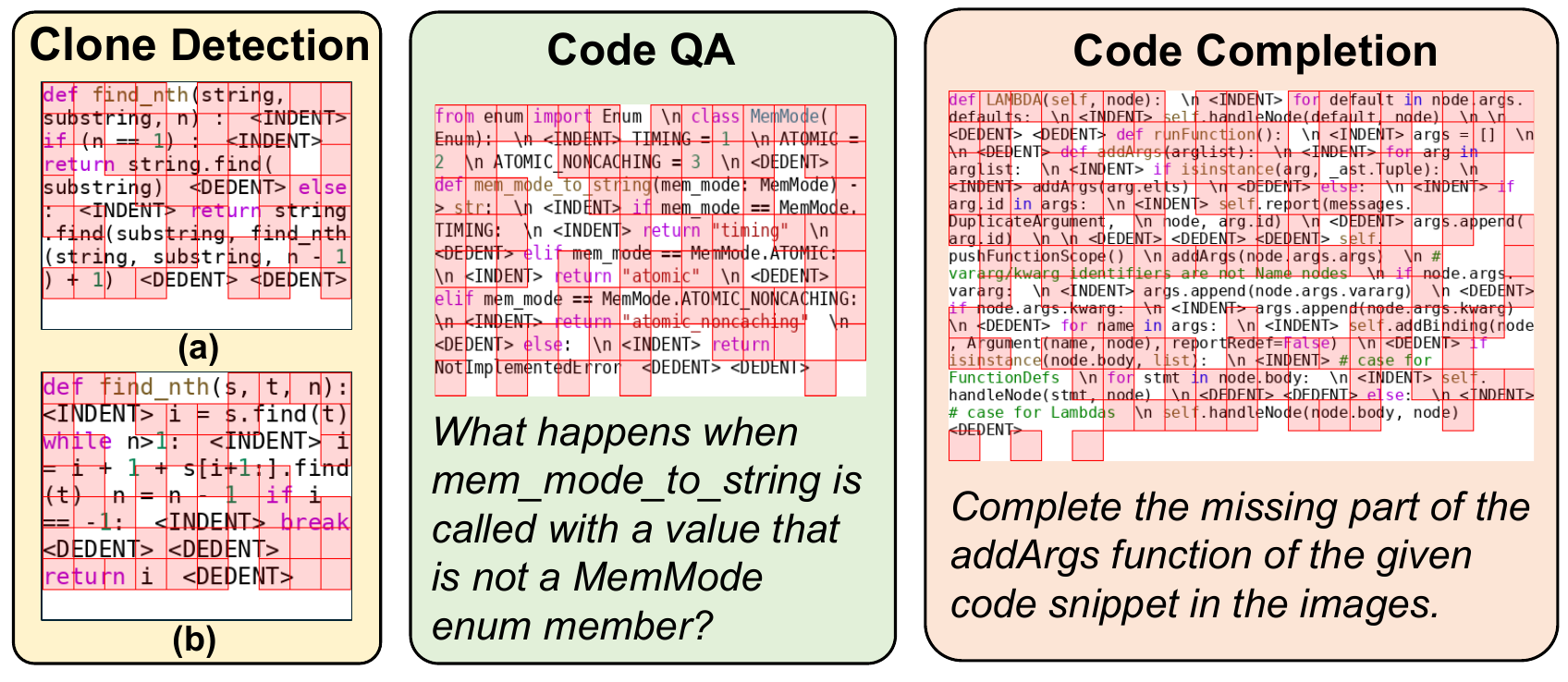}
  \caption{Case study of the DTS on the three tasks, where the red boxes mark the visual tokens retained by DTS. Code clone detection contains two snippets, and the instruction is shown below.}

  \label{fig:case_prune}
\end{figure}

As shown in Fig.~\ref{fig:case_prune}, the tokens retained by DTS align closely with the semantic need of each task rather than being uniformly distributed. For Code QA, where the question concerns the behavior of a specific function, the model concentrates its attention on the conditional branches and traces them step by step until reaching the else branch that determines the answer. For Code Clone Detection, where the goal is to decide whether two snippets are equivalent, the model attends to roughly symmetric regions across the two images, comparing their corresponding structures instead of focusing on either one alone. These patterns indicate that the retained tokens are exactly the ones the task relies on.

\subsubsection{\aps} For ease of illustration, here we only show the choices of the scale ratio. As shown in Fig.~\ref{fig:case_scale}, the Config Agent selects 4$\times$ for this sample. A smaller ratio such as 1$\times$ or 2$\times$ keeps an unnecessarily high resolution and wastes visual tokens, whereas a larger ratio such as 8$\times$ shrinks the font to the point where the code becomes illegible to the model. By choosing 4$\times$, the Agent reaches the best balance between legibility and compression, which illustrates how it adapts the configuration to the specific content of each sample.

\begin{observationbox}
\textbf{Answer to RQ5:} \method\ works through three complementary mechanisms: (1) BFR removes blank redundancy to produce dense and legible code images, (2) DTS retains the instruction-relevant tokens for each task, and (3) ACC selects a suitable compression configuration for each sample, so that aggressive compression still preserves the information needed by the task.
\end{observationbox}

\section{Related Work}

\subsection{Multimodal Large Language Models for Code}
The integration of visual modalities into software engineering has attracted increasing attention. Existing work mainly falls into two categories: visual code understanding benchmarks~\cite{tang2025slidecoder,li2024mmcode,zhang2025humanevalV,tang2026efficientpostergen}, which evaluate MLLMs' capabilities on multimodal coding tasks, and UI-to-code generation~\cite{Dang2025Envisioning,xiao2026comuicoder,le2026uibenchkit}, which generates source code from UI screenshots or design mockups~\cite{wan2026runnable, wan2024mrweb, xiao2024interaction2code,yuan2025designrepair, xiao2025designbench}. Beyond UI generation, AgileGen~\cite{Agile} incorporates visual artifacts into AI-assisted software development through human-AI collaboration. Unlike these studies, \method\ treats source code itself as the visual input and addresses layout-induced token redundancy and adaptive visual compression for efficient code understanding.

\subsection{Token Compression}


(1) \textbf{Text-based Token Compression.} Early efforts to reduce input cost operate directly in the textual token space. LLMLingua~\cite{jiang2023llmlingua} identifies and removes non-essential tokens, and is further extended by LongLLMLingua~\cite{jiang2024longllmlingua} for question-aware long-context compression and LLMLingua-2~\cite{pan2024llmlingua2} through token-level compression via knowledge distillation. For programming tasks, LongCodeZip~\cite{shi2025LongCodeZip} proposes a dual-stage code compression framework with adaptive token budgeting. Recent studies further improve code token efficiency by introducing semantics-preserving shorthand representations~\cite{sun2025token}, compressing docstrings while maintaining code generation performance~\cite{yang2026less}, and reducing unnecessary formatting overhead in source code~\cite{pan2025hidden}. (2)\textbf{Visual Token Compression.} For MLLMs, visual token compression has emerged as a complementary direction. FastV~\cite{chen2024fastv} is the first to identify inefficient visual attention in MLLMs, proposing a training-free method that prunes tokens with the lowest visual-text attention scores after an early transformer layer. VisionZip~\cite{yang2025visionzip} selects visual tokens based on visual encoder attention and applies a merging strategy to retain residual information. For code-related tasks, EfficientUICoder~\cite{xiao2025efficientuicoder} addresses token redundancy in UI2Code tasks via element-aware compression combined with attention-based token refinement. CodeOCR~\cite{shi2026codeocr} explores representing source code as rendered images, showing that MLLMs can effectively understand code under substantial resolution-based compression. However, these methods either target natural images or UI screenshots without addressing code-specific layout redundancy, and rely on fixed scale ratios that do not adapt to task or model preferences. \method \ addresses these limitations by eliminating layout-induced blank tokens , pruning task-irrelevant tokens, and learning a adaptive compression configuration.

\section{Threats to Validity}

(1) We evaluate \method \ on three representative code understanding tasks using widely adopted benchmarks and standard metrics. While these tasks cover a broad range of scenarios, evaluating \method \ on additional software engineering tasks could further demonstrate its applicability.  (2) Our experiments are conducted on Python and Java, which represent two widely used programming languages. Extending the evaluation to more programming languages and larger real-world software repositories would provide additional evidence of the generalizability of \method.

\section{Conclusion}

In this paper, we investigated visual token redundancy in code images for MLLM-based code understanding, finding that blank regions, task-irrelevant code, and fixed compression configurations lead to inefficient visual token usage. We proposed \method, an adaptive visual compression framework consisting of \bfr, which removes layout-induced blank tokens, \iap, which prunes instruction-irrelevant visual tokens, and \aps, which predicts per-input compression configurations via reinforcement learning. Experiments on code question answering, clone detection, and code completion show that \method \ substantially reduces visual tokens while matching or even surpassing uncompressed text-only inputs, consistently outperforming both text-based and visual compression baselines.



\bibliographystyle{IEEEtran} 
\bibliography{ref}

\end{document}